\documentclass[draft]{article}

\usepackage[T1]{fontenc}
\usepackage[latin1]{inputenc}
\usepackage{amsmath,amsfonts,amssymb,url}
\usepackage{url}
\usepackage[final]{graphicx}
\usepackage{ijfcs}

\def\ra{\rightarrow}
\def\Ra{\Rightarrow}
\def\se{\subseteq}
\def\es{\emptyset}

\def\lH{\mathrm{lH}}
\def\nH{\mathrm{cH}}

\def\ssH{\mathrm{ssH}}
\def\aH{\mathrm{aH}}
\def\qH{\mathrm{qaH}}
\def\saH{\mathrm{saH}}
\def\lcH{\mathrm{lcH}}

\def\vp{\varphi}
\def\calB{\mathcal{B}}
\def\Rec{\mathrm{Rec}}

\DeclareMathOperator{\yd}{yd}
\DeclareMathOperator{\rk}{rk}
\DeclareMathOperator{\var}{var}
\DeclareMathOperator{\pos}{pos}
\DeclareMathOperator{\id}{id}
\DeclareMathOperator{\br}{br}
\DeclareMathOperator{\sub}{sub}
\DeclareMathOperator{\card}{card}
 \DeclareMathOperator{\hg}{hg}

\providecommand*{\seq}[3]{\ensuremath{#1_{#2}, \dotsc,
#1_{#3}}}
\providecommand*{\word}[3]{\ensuremath{#1_{#2} \dotsm #1_{#3}}}
\providecommand*{\nat}[0]{\ensuremath{\mathbb{N}}}

\begin{document}
\copyrightheading
\symbolfootnote
\textlineskip

\begin{center}

\fcstitle{PROPERTIES OF QUASI-ALPHABETIC TREE BIMORPHISMS}

\vspace{24pt}

{\authorfont Andreas Maletti and C\u at\u alin Ionu\c t T\^{i}rn\u
  auc\u a}

\vspace{2pt}

\smalllineskip
{\addressfont Universitat Rovira i Virgili \\
  Pla\c ca Imperial T\`arraco 1, 43005 Tarragona, Spain \\[1ex]
  \begin{tabular}{ll}
    email: & \url{andreas.maletti@urv.cat} \\
    & \url{catalinionut.tirnauca@estudiants.urv.cat}
  \end{tabular}}

\vspace{20pt}

\publisher{(received date)}{(revised date)}{Editor's name}

\end{center}

\alphfootnote

\begin{abstract}
  We study the class of quasi-alphabetic relations, i.e., tree
  transformations defined by tree bimorphisms $(\vp,L,\psi)$ with
  $\vp,\psi$ quasi-alphabetic tree homomorphisms and $L$ a regular
  tree language. We present a canonical representation of these
  relations; as an immediate consequence, we get the closure under
  union.  Also, we show that they are not closed under intersection
  and complement, and do not preserve most common operations on trees
  (branches, subtrees, $v$-product, $v$-quotient,
  $f$-top-catenation). Moreover, we prove that the translations
  defined by quasi-alphabetic tree bimorphism are exactly products of
  context-free string languages. We conclude by presenting the
  connections between quasi-alphabetic relations, alphabetic relations
  and classes of tree transformations defined by several types of
  top-down tree transducers. Furthermore, we get that quasi-alphabetic
  relations preserve the recognizable and algebraic tree languages.

\end{abstract}

\textlineskip

\section{Introduction}
\label{sec:Intro}
Tree transformations were extensively study in the past four decades
from the algebraic point of view offered by tree bimorphisms
\cite{arndau82,Bozapalidis92,ste86,tak72,tak77} or from the dynamic
point of view provided by tree transducers
\cite{bak78,eng75,gecste84,rah01,rou70}. Recently, new types of tree
transducers were used with considerable success in modeling
translations between natural languages especially because of their
ability to capture syntax-sensitive transformations and to do certain
reorderings of parts of sentences. This way, the new field of
syntax-based machine translation was established (see
\cite{graknimay08,kni07,knigra05,malgrahopkni07} and the references
therein). Unfortunately, properties that may improve the translation
process (e.g., closure under composition and preservation of
recognizable and algebraic tree languages~\cite{kni07,knigra05}) do
not hold in general for most of the main tree transducer types
\cite{bak79,eng75,gecste84,malgrahopkni07}.

First proposed as models of a compiler \cite{Irons61}, synchronous
grammars represent classes of tree transformations that describe in a
natural way translations between natural languages
\cite{ahoull72,satpes05,shi04,shisch90}. It consists of two formal
grammars with productions linked by some criteria, pairs of
recursively related sentences being generated simultaneously. This
way, not only that they model the syntax-sensitive transformations
between natural languages, but moreover, they intrinsically describe
(perform) difficult local rotations required by natural language pairs
with extremely different structures such as Arabic-English or
Chinese-English. Unfortunately, the mathematical framework offered by
such formalisms is quite poor since for example, no closure results
were known \cite{shi04}.

An elegant algebraic way to define tree transformations is by the tree
bimorphism formalism which is formed by two tree homomorphisms defined
on the same common tree language. Tree bimorphisms were used with
considerable success in proving properties like closure under
composition and preservation of recognizability, especially when
suitable restrictions were imposed on its constituents
\cite{arndau82,Bozapalidis92,ste86,tak72,tak77}. Moreover, by taking
the yields of the input trees and output trees, they are transformed
into word-for-word translation devices. A survey on the main classes
of tree bimorphisms and their characteristics is \cite{tir08a}.

Using the tree bimorphism formalism, \textsc{S.M. Shieber} was the
first one who linked tree transducers and synchronous grammars in an
attempt to improve the mathematical framework of the later devices
\cite[p.95]{shi04}: ''...the \emph{bimorphism characterization} of
tree transducers has led to a series of composition closure results.
Similar techniques may now be \emph{applicable} to synchronous
formalisms, \emph{where no composition results are known}...''
Following this lead, the class of quasi-alphabetic tree bimorphisms
that define the same translations as syntax-directed translation
schemata of \cite{ahoull72} was introduced in \cite{stetir07}. In
\cite{tir08b} connections between these tree bimorphisms and other
synchronous grammars are presented in detail, and similar results
involving other types of tree bimorphisms are summed up in
\cite{tir08a}.

It was already shown in \cite{stetir07} that the tree transformations
defined by quasi-alphabetic tree bimorphisms, called
\emph{quasi-alphabetic relations} here, are closed under composition
and inverses, and preserve the recognizability of tree languages. In
the present work we further investigate the properties of this class
from a theoretical point of view (What other closure properties has or
what common operations on trees are preserved? Is there any canonical
representation of such a class? What is its place in the tree
transducer hierarchy?) but also having in mind their connection with
synchronous grammars (What other mathematical properties can be
transferred to several types of synchronous grammars? What is the
power of the translations defined by quasi-alphabetic bimorphisms?).

Our results can be summarized as follows. After presenting in
Section~\ref{sec:Prelims} the basic definitions and notions used, we
show in Section~\ref{sec:Quasi} that there is a canonical
representation of quasi-alphabetic relations, and
by using it, that these relations are closed under union. Also, we
show that they are not closed under intersection and complement. After
this, we turn our attention to what common operation on trees are
preserved by such relations: we found out that intersection and
reunion with a regular language are preserved, but branches, subtrees,
$v$-product, $v$-quotient, $f$-top-catenation are not, in general. We
end Section~\ref{sec:Quasi} by proving a more general result of
\cite{stetir07}: the translations defined by quasi-alphabetic tree
bimorphisms are exactly
the products of context-free string
languages. Section~\ref{sec:Relation} is dedicated
to the connection of quasi-alphabetic relations with other well-known
classes of tree transformations: alphabetic transductions
\cite{Bozapalidis92}, finite-state relabelings \cite{eng75}, tree
transformations defined by several types of top-down tree transducers
\cite{eng75} and top-down tree transducers with look-ahead
\cite{eng77}. All the results are depicted in the Hasse diagram of
Figure~\ref{fig:Hasse}. Moreover, as an immediate consequence of the
fact that quasi-alphabetic relations are strictly included in the
class of alphabetic ones, we get that our class preserves recognizable
and algebraic tree languages, too.

\section{Preliminaries}
\label{sec:Prelims}
Let $R$, $S$, and~$T$ be sets, and consider a relation $\tau \se
S\times T$. The fact that $(s,t)\in \tau$ can also be expressed by
writing $s \mathrel\tau t$. For every $s\in S$, let $s\tau = \{ t \mid
s \mathrel\tau t \}$.  More generally, for every $A \subseteq S$, we
let $A\tau = \bigcup_{a \in A} a\tau$. The \emph{inverse} of $\tau$ is
the relation $\tau^{-1}=\{ (t,s) \mid s \mathrel\tau t \}$. The
\emph{composition} of two relations $\rho \subseteq R \times S$ and
$\tau \subseteq S \times T$ is the relation $\rho \circ \tau = \{
(r,t) \mid \exists s\in S \colon r \mathrel\rho s \mathrel\tau t
\}$. The \emph{identity relation}~$\id_S$ is $\{(s,s) \mid s \in
S\}$. If~$S$ is understood, then we simply write~$\id$. For (total)
mappings $\vp \colon S \ra T$ we generally identify
$s\vp$~and~$\vp(s)$ for every $s\in S$. The nonnegative integers are
denoted by~$\nat$. For every $k \in \nat$, the set $\{i \in \nat \mid
1 \leq i \leq k\}$ is denoted by~$[k]$.

For a set~$V$, we denote by $V^*$ the set of \emph{strings} over $V$,
and $\varepsilon$ denotes the \emph{empty string}. By an \emph{alphabet}
we mean a finite set of symbols.  A \emph{ranked alphabet} $(\Sigma,
\rk)$ consists of an alphabet~$\Sigma$ and a mapping~$\rk \colon
\Sigma \to \nat$. Often we leave the mapping~$\rk$ implicit. For every
$k \geq 0$, let $\Sigma_k = \{ f \in \Sigma \mid \rk(f) = k\}$. We may
write $\Sigma = \{ f_1/k_1,\dotsc,f_n/k_n \}$ to indicate that
$\Sigma$ consists of the symbols $\seq f1n$ with the respective ranks
$\seq k1n$.

Let $\Sigma$ be a ranked alphabet and $T$ a set. Then
\[ \Sigma(T) = \{ f(\seq t1k) \mid f \in \Sigma_k, \seq t1k \in T\}
\enspace. \] For every (leaf) alphabet~$V$, the set~$T_\Sigma(V)$ of
all $\Sigma$-\emph{trees indexed by}~$V$ is the smallest set~$T$ such
that $V \subseteq T$ and $\Sigma(T) \subseteq T$. Subsets of
$T_\Sigma(V)$ are called (\emph{tree}) \emph{languages}. Generally,
for all considered trees we assume that the ranked alphabet is
disjoint with the leaf alphabet. For every tree~$t \in T_\Sigma(V)$,
the set~$\pos(t) \subseteq \nat^*$ of \emph{positions} of~$t$ is
inductively given by $\pos(v) = \{\varepsilon\}$ for every $v \in V$,
and
\[ \pos(f(\seq t1k)) = \{\varepsilon\} \cup \{ iw \mid i \in [k], w
\in \pos(t_i)\} \] for every $f \in \Sigma_k$ and $\seq t1k \in
T_\Sigma(V)$. The \emph{label} of~$t$ at position~$w \in \pos(t)$ is
denoted by~$t(w)$, the \emph{subtree} of~$t$ at~$w$ is denoted
by~$t|_w$, and the replacement of that subtree in~$t$ by the tree~$u
\in T_\Sigma(V)$ is denoted by~$t[u]_w$. For every $\Omega \subseteq
\Sigma \cup V$ let $\pos_\Omega(t) = \{ w \in \pos(t) \mid t(w) \in
\Omega\}$ and $\pos_f(t) = \pos_{\{f\}}(t)$ for every $f \in \Sigma
\cup V$. The set of \emph{branches} of t is $\br(t) = \pos_{\Sigma_0
  \cup V}(t)$, and the set of subtrees of $t$ is~$\sub(t)= \{ t|_w
\mid w \in \pos(t) \}$. Finally, $\lvert t \rvert_f =
\card(\pos_f(t))$, and the \emph{height}~$\hg(t)$ is the length of a
longest string in~$\pos(t)$.

A tree~$t \in T_\Sigma(V)$ is \emph{linear} (respectively,
\emph{nondeleting}) in~$Y \subseteq V$ if $\lvert t \rvert_y \leq 1$
(respectively, $\lvert t \rvert_y \geq 1$) for every $y \in Y$. The
$Y$-\emph{yield} of a tree~$t \in T_\Sigma(V)$ is defined inductively
by $\yd_Y(y) = y$ for every $y \in Y$, $\yd_Y(v) = \varepsilon$ for
every $v \in V \setminus Y$, and $\yd_Y(f(\seq t1k)) = \yd_Y(t_1)
\dotsm \yd_Y(t_k)$ for every $f \in \Sigma_k$ and $\seq t1k \in
T_\Sigma(V)$.

We fix a set~$X = \{x_i \mid i \geq 1\}$ of \emph{formal variables}
(disjoint to all other ranked alphabets and leaf alphabets). Let $n
\geq 0$. We let $X_n = \{x_i \mid i \in [n]\}$ and
\[ C_\Sigma^n(V) = \{ t \in T_\Sigma(V \cup X_n) \mid \forall i \in
[n] \colon \lvert t \rvert_{x_i} = 1 \} \enspace. \] In particular,
the elements of $C_\Sigma^1(V)$ are called \emph{contexts}. For every
$t \in T_\Sigma(V \cup X_n)$, let $\var(t) = \{ x \in X_n \mid
\pos_x(t) \neq \emptyset\}$.

For all $t, \seq t1n \in T_\Sigma(V \cup X_n)$, we denote by~$t[\seq t1n]$ the result obtained by replacing, for
every~$i \in [n]$, every occurrence of~$x_i$ in~$t$ by~$t_i$. For all $L, \seq L1n \subseteq T_\Sigma(V \cup
X_n)$, $L[\seq L1n]$ denotes $\bigcup_{t \in L,
  t_1 \in L_1, \dotsc, t_n \in L_n} t[\seq t1n]$. Let $n = \lvert t
\rvert_v$. More generally, for every $v \in V$, the result of
replacing, for every $i \in [n]$, the $i$-th (with respect to the
usual lexicographic order on the positions) occurrence of~$v$ by~$t_i$
is denoted by $t[v \gets (\seq t1n)]$.  For every $t \in T_\Sigma(V
\cup X_n)$, $f \in \Sigma_1$, and $c \in C_\Sigma^1(V)$, we let
$f^0(t) = t$~and~$C^0=C$, and $f^{k+1}(t) = f(f^k(t))$ and $C^{k+1} =
C[C^k]$ for all $k \geq 0$.

For every $f \in \Sigma_k$ and $\seq L1k \subseteq T_\Sigma(V \cup X_n)$, the \emph{$f$-top-catenation} of $\seq
L1k$ is $f(\seq L1k) = \bigcup_{t_1 \in L_1, \dotsc, t_k \in L_k} f(\seq t1k)$.  Moreover for every $v \in V$,
the $v$-\emph{product}~$L \bullet_v L'$ of two languages $L, L' \se T_\Sigma(V)$ is
\[ L \bullet_v L' = \bigcup_{\substack{t \in L, n = \lvert t \rvert_v
    \\ \forall i \in [n] \colon t_i \in L'}} t[v \gets (\seq t1n)]
\enspace. \] Then, the $v$-\emph{quotient} of~$L$ by~$L'$ is $L /_v L'
= \{ t\in T_\Sigma(V) \mid \{ t \} \bullet_v L' \cap L \neq \es
\}$. For a more detailed description of those operations on tree
languages, we refer the reader to~\cite{Bozapalidis92}.

A (\emph{tree}) \emph{homomorphism} $\vp \colon T_\Sigma(V) \ra
T_\Delta(Y)$ can be presented by a mapping $\vp_V \colon
V \ra T_\Delta(Y)$ and mappings $\vp_k \colon \Sigma_k \ra T_\Delta(Y \cup X_k)$ for every $k\geq 0$ as follows:
\begin{itemize}
\item[(i)] $v\vp = \vp_V(v)$ for every $v\in V$, and
\item[(ii)] $f(\seq t1k) \vp = \vp_k(f) [t_1\vp, \dotsc, t_k\vp]$ for
  every $f \in \Sigma_k$ and $\seq t1k \in T_\Sigma(V)$.
\end{itemize}
We say that it is \emph{normalized} if for every $f \in \Sigma_k$ there exists $n \geq 0$ such that
$\yd_X(\vp_k(f)) = x_1 \dots x_n$. Moreover, such a homomorphism $\vp$ is
\begin{itemize}
\item \emph{linear}~\cite{gecste84,Bozapalidis92,tata97}
  (respectively, \emph{complete}~\cite{tata97}) if $\vp_k(f)$ is
  linear (respectively, nondeleting) in~$X_k$ for every $f \in
  \Sigma_k$,
\item \emph{symbol-to-symbol}~\cite{tata97} if $\vp_V(v)\in Y$ for
  every $v \in V$ and $\vp_k(f) \in \Delta(X_k)$ for every $f \in
  \Sigma_k$,
\item \emph{alphabetic}~\cite{Bozapalidis92,arndau78} (d\'emarquage
  lin\'eaire in~\cite{arndau78}) if it is linear, $\vp_V(v)\in Y$ for
  every $v\in V$, and $\vp_k(f) \in X_k \cup \Delta(X_k)$ for every $f
  \in \Sigma_k$, and
\item \emph{strictly alphabetic}~\cite{Bozapalidis92} if it is
  complete, alphabetic and symbol-to-symbol.
\end{itemize}

We denote by $\lH$, $\nH$, $\ssH$, $\aH$, and $\saH$ the classes of
all linear, complete, symbol-to-symbol, alphabetic, and strictly
alphabetic tree homomorphisms, respectively. Further subclasses of
tree homomorphisms can be obtained by combining any of these
restrictions. For example, $\lcH$ is the class of all linear complete
tree homomorphisms.

A (\emph{tree}) \emph{bimorphism} is a triple $B=(\vp, L, \psi)$ where
$L \se T_\Gamma(Z)$ is a tree language, $\vp \colon T_\Gamma(Z) \ra
T_\Sigma(V)$, and $\psi \colon T_\Gamma(Z)\ra T_\Delta(Y)$ are
homomorphisms. The \emph{tree transformation defined by}~$B$ is
\[ \tau_B=\vp^{-1} \circ \mathord{\id_L} \circ \psi=\{ (t\vp, t\psi)
\mid t \in L \} \enspace. \] The \emph{translation defined by}~$B$ is
\[ \yd(\tau_B) = \{(\yd_V(t\vp),\yd_Y(t\psi)) \mid t \in L \} = \{
(\yd_V(t), \yd_Y(u)) \mid (t,u) \in \tau_B\} \enspace. \] For all
classes ${\cal H}_1$ and ${\cal H}_2$ of homomorphisms and every class
$\cal L$ of tree languages, we denote by $\calB({\cal H}_1,{\cal
  L},{\cal H}_2)$ the class of tree transformations~$\tau_B$ where
$B=(\vp,L,\psi)$ with $\vp \in {\cal H}_1$, $L \in \cal L$ and
$\psi\in {\cal H}_2$.

A \emph{top-down tree transducer}~\cite{rou70,tha70} is a tuple $M =
(Q, \Sigma, \Delta, I, R)$ where
\begin{itemize}
\item $Q=Q_1$ is a unary ranked alphabet of~\emph{states} disjoint
  with $\Sigma \cup \Delta$,
\item $\Sigma$~and~$\Delta$ are an \emph{input} and an \emph{output
    alphabet}, respectively,
\item $I \subseteq Q$ is a set of \emph{final states}, and
\item $R$ is a finite set of \emph{rules} of the form $l \to r$ where
  $l \in Q(T_\Sigma(X))$ is linear in~$X$ and $r \in
  T_\Delta(Q(\var(l)))$.
\end{itemize}
The top-down tree transducer $M = (Q, \Sigma, \Delta, I, R)$ is
\emph{linear} (respectively, \emph{nondeleting}) if $r$ is linear
(respectively, nondeleting) in~$\var(l)$ for every rule $l \to r \in
R$. The one-step \emph{derivation relation} $\Ra_M$ is defined as
follows. For every $s, t\in T_\Delta(Q(T_\Sigma))$ we have $s \Ra_M t$
if and only if there exists a rule $l \to r \in R$, a position~$w \in
\pos(s)$, and $\seq u1n \in T_\Delta$ where $n = \rk(s(w))$ such that
$s|_w = l[\seq u1n]$ and $t = s[u]_w$ with $u = r[\seq u1n]$. Let
$\Ra_M^*$ be the reflexive and transitive closure of~$\Ra_M$. The
\emph{tree transformation computed by}~$M$ is
\[ \tau_M = \{ (s,t) \in T_\Sigma \times T_\Delta \mid \exists q \in I
\colon q(s) \Ra_M^* t \} \enspace. \] The class of all tree
transformations computable by linear (respectively, linear and
nondeleting) top-down tree transducers is denoted by $\text{l-TOP}$
(respectively, $\text{ln-TOP}$).

Let $M = (Q, \Sigma, \Delta, I, R)$ be a top-down tree transducer. It
is a \emph{finite-state relabeling}~\cite{eng75}, if every rule $l \to
r \in R$ is of the form $l = q(f(\seq x1k))$ and $r = g(q_1(x_1),
\dotsc, q_k(x_k))$ for some $q, \seq q1k \in Q$, $f \in \Sigma_k$, and
$g \in \Delta_k$. If additionally, $l(1) = r(\varepsilon)$ for every
$l \to r \in R$, then $M$ is a \emph{finite-state tree automaton}
(fta)~\cite{eng75}.  We generally write rules of an fta in the form $q
\to f(\seq q1k)$ instead of $q(f(\seq x1k)) \to f(q_1(x_1), \dotsc,
q_k(x_k))$. Note that $\tau_M$ coincides with~$\id_L$ for some $L
\subseteq T_\Sigma$, if $M$ is an fta.  This $L$ is also denoted
by~$L(M)$, and additionally, for every $q \in Q$, the
notation~$L(M)_q$ stands for $L(N)$ where $N = (Q, \Sigma, \Delta,
\{q\}, R)$. A language~$L$ is \emph{recognizable} if there exists an
fta~$N$ such that $L(N) = L$. The class of recognizable tree
languages~\cite[Chapter~II]{gecste84} is denoted by~$\Rec$. Finally,
$M$ is a \emph{relabeling}~\cite{eng75} if it a finite-state
relabeling and $\card(Q) = 1$.  We denote the classes of
transformations computed by finite-state relabelings, relabelings, and
fta by $\text{\rm QREL}$, $\text{\rm REL}$, and $\text{\rm FTA}$,
respectively.

The top-down tree transducer~$M$ can be equipped with a look-ahead
facility~\cite{eng77,malgrahopkni07}. The pair $\langle M, c\rangle$
where $M = (Q, \Sigma, \Delta, I, R)$ is a top-down tree transducer
and $c \colon R \to {\cal P}(T_\Sigma)$ is called a \emph{top-down
  tree transducer with look-ahead}. The look-ahead~$c$ is
\emph{regular} (or \emph{recognizable}), if $c(l \to r)$ is
recognizable for every $l \to r \in R$, and it is \emph{finite}, if
$c(l \to r) \in L[T_\Sigma, \dotsc, T_\Sigma]$ for a finite tree
language~$L \subseteq T_\Sigma(X)$. In the latter case, we often write
$c(l \to r) = L$. The transducer $\langle M, c\rangle$ inherits the
properties `linear' and `nondeleting' from~$M$. The semantics of a
top-down tree transducer~$\langle M, c\rangle$ with look-ahead is
defined as for the top-down tree transducer~$M$ with the additional
condition that $s|_{w} \in c(l \to r)$ in the definition
of~$\Ra_M$. The class of transformations computed by linear top-down
tree transducers with finite (respectively, regular) look-ahead is
denoted by~$\text{\rm l-TOP}^{\text{\rm F}}$ (respectively, $\text{\rm
  l-TOP}^{\text{\rm R}}$).

\section{Properties of Quasi-Alphabetic Relations}
\label{sec:Quasi}
Let us start by recalling the main notion of this contribution. A
quasi-alphabetic homomorphism is linear, complete, and basically
symbol-to-symbol, but allows variables as successors of an output
symbol. The precise definition follows.

\begin{definition}[{\protect{see~\cite[Section~3]{stetir07}}}]
  \label{df:Quasi}
  A tree homomorphism $\vp \colon T_\Sigma(V) \ra T_\Delta(Y)$ is
  \emph{quasi-alphabetic} if
  \begin{itemize}
  \item[(i)] it is linear and complete,
  \item[(ii)] $\vp_V(v)\in Y$ for every $v\in V$, and
  \item[(iii)] $\vp_k(f) \in \Delta(Y \cup X_k)$ for every $f \in
    \Sigma_k$.
  \end{itemize}
  By~$\qH$ we denote the class of all quasi-alphabetic
  homomorphisms. A \emph{quasi-alphabetic bimorphism} is a
  bimorphism~$(\vp, L, \psi)$ such that $\vp$~and~$\psi$ are
  quasi-alphabetic and $L$ is recognizable.
\end{definition}

The name `quasi-alphabetic' deserves some discussion. They are called
such because they are almost `alphabetic' in the sense
of~\cite{gecste84} (a relabeling in our terminology). Note that we
here use the notion of `alphabetic' that is used
in~\cite{Bozapalidis92}. Thus, with our terminology in mind, we might
have called them `quasi-relabelings'. However, the term
`quasi-alphabetic' is established~\cite{stetir07} and we continue to
use it. In particular, ${\cal B}(\qH,\Rec,\qH)$ is the class of all
the tree transformations defined by quasi-alphabetic bimorphisms; such
relations are called \emph{quasi-alphabetic relations}.

Every quasi-alphabetic homomorphism maps each input symbol to an
output symbol possibly with some output leaf variables as direct
subtrees. However, the variables of~$X$ have to occur as direct
subtrees of the root output symbol. This immediately yields the
following proposition.

\begin{proposition}
  \label{prop:Height}
  Let $\vp :T_\Sigma(V) \ra T_\Delta(Y)$ be a homomorphism and $t\in
  T_\Sigma(V)$.
  \begin{itemize}
  \item If $\vp$ is quasi-alphabetic, then $\hg(t) \leq \hg(t\vp) \leq
    \hg(t)+1$.
  \item If $\vp$ is symbol-to-symbol, then $\hg(t\vp) \leq \hg(t)$.
  \item If $\vp$ is strictly alphabetic, then $\hg(t\vp) = \hg(t)$.
  \end{itemize}
\end{proposition}

Now we investigate the fundamental properties of quasi-alphabetic
relations. We start our investigation with a canonical representation
of quasi-alphabetic relations in the spirit
of~\cite[Proposition~3.1]{Bozapalidis92}. This representation will
allow us to conclude that quasi-alphabetic relations are closed under
union.

For the rest of this section, let $B = (\vp, L, \psi)$ with $\vp
\colon T_\Gamma(Z) \ra T_\Sigma(V)$ and $\psi \colon T_\Gamma(Z) \ra
T_\Delta(Y)$ be a quasi-alphabetic bimorphism. Let $[\Sigma \times
\Delta]$ be the ranked alphabet such that for every $k \geq 0$
\[ [\Sigma \times \Delta]_k = \{\langle t, u \rangle \mid t \in
\Sigma(V \cup X_k) \cap C^k_\Sigma(V), u \in \Delta(Y \cup X_k) \cap
C^k_\Delta(Y) \} \enspace. \] There are canonical quasi-alphabetic
homomorphisms $\rho^1 \colon T_{[\Sigma \times \Delta]}(V \times Y)
\to T_\Sigma(V)$ and $\rho^2 \colon T_{[\Sigma \times \Delta]}(V
\times Y) \to T_\Delta(Y)$ given by
\begin{align*}
  \rho^1_{V \times Y}(\langle v,y \rangle) &= v & \rho^1_k(\langle t,
  u \rangle) &= t \\
  \rho^2_{V \times Y}(\langle v,y \rangle) &= y & \rho^2_k(\langle t,
  u \rangle) &= u
\end{align*}
for every $\langle v, y \rangle \in V \times Y$ and $\langle t, u
\rangle \in [\Sigma \times \Delta]_k$. Henceforth, we will use these
projections also for other product ranked alphabets.

\begin{proposition}[{\protect{see~\cite[Proposition~3.1]{Bozapalidis92}}}]
  \label{prop:Rep}
  There exists a quasi-alphabetic homomorphism $\eta \colon
  T_\Gamma(Z) \to T_{[\Sigma \times \Delta]}(V \times Y)$ such that
  $t\varphi = (t \eta) \rho^1$ and $t\psi = (t \eta) \rho^2$ for every
  $t \in T_\Gamma(Z)$.
\end{proposition}
\proof{
  Let $\eta \colon T_\Gamma(Z) \to T_{[\Sigma \times \Delta]}(V \times
  Y)$ be the tree homomorphism such that $\eta_Z(z) = \langle \varphi_Z(z),
  \psi_Z(z) \rangle$ for every $z \in Z$ and $\eta_k(f) = \langle
  \varphi_k(f), \psi_k(f) \rangle$ for every $f \in
  \Gamma_k$. Clearly, $\eta$ is quasi-alphabetic, and it is easy to
  check that $t\varphi = (t \eta) \rho^1$ and $t\psi = (t \eta)
  \rho^2$ for every $t \in T_\Gamma(Z)$.
}

Using the previous proposition, we can now eliminate from $B$ the
ranked alphabet~$\Gamma$, the index set~$Z$, and the particular tree
homomorphisms $\vp$~and~$\psi$. Essentially, every quasi-alphabetic
relation $\tau \subseteq T_\Sigma(V) \times T_\Delta(Y)$ is determined
by a recognizable language $L \subseteq T_{[\Sigma \times \Delta]}(V
\times Y)$.

\begin{theorem}[{\protect{see~\cite[Proposition~3.1]{Bozapalidis92}}}]
  \label{thm:Fact}
  A relation $\tau \subseteq T_\Sigma(V) \times T_\Delta(Y)$ is
  quasi-alphabetic if and only if there exists a recognizable language
  $L \subseteq T_{[\Sigma \times \Delta]}(V \times Y)$ such that $\tau
  = \{ (t\rho^1, t\rho^2 \mid t \in L\}$.
\end{theorem}
\proof{
  The if-direction is trivial since $(\rho^1, L, \rho^2)$ is a
  quasi-alphabetic bimorphism defining~$\tau$. For the converse, let
  $B = (\vp, L', \psi)$ be a quasi-alphabetic bimorphism such that
  $\tau_B = \tau$. By Proposition~\ref{prop:Rep} there exists a
  quasi-alphabetic homomorphism $\eta \colon T_\Gamma(Z) \to
  T_{[\Sigma \times \Delta]}(V \times Y)$ such that $\tau = \{
  (t\eta\rho^1, t\eta\rho^2) \mid t \in L'\}$. Consequently, the
  language $L = \eta(L')$ has the desired properties because it is
  recognizable by~\cite[Theorem~II.4.16]{gecste84}.
}

We immediately note that quasi-alphabetic relations are trivially
closed under inverses~\cite[Theorem~4]{stetir07}; i.e., if $\tau \in
{\cal B}(\qH, \Rec, \qH)$, then so is~$\tau^{-1}$.  As promised, let
us use the previous theorem to prove that quasi-alphabetic relations
are closed under union.

\begin{corollary}[{\protect{cf.~\cite[Proposition~3.2]{Bozapalidis92}}}]
  \label{cor:Union}
  ${\cal B}(\qH, \Rec, \qH)$ is closed under union.
\end{corollary}
\proof{
  Let $\tau_1, \tau_2 \subseteq T_\Sigma(V) \times T_\Delta(Y)$ be
  quasi-alphabetic relations. By Theorem~\ref{thm:Fact}, there exist
  recognizable $L_1, L_2 \subseteq T_{[\Sigma \times \Omega]}(V \times
  Y)$ such that
  \[ \tau_1 = \{ (t\rho^1, t\rho^2) \mid t \in L_1\} \quad \text{ and
  } \quad \tau_2 = \{ (t\rho^1, t\rho^2) \mid t \in L_2\} \enspace. \]
  Then
  \begin{align*}
    \tau_1 \cup \tau_2 &= \{ (t\rho^1, t \rho^2) \mid t \in L_1 \}
    \cup \{ (t \rho^1, t \rho^2) \mid t \in L_2\} = \{ (t \rho^1, t
    \rho^2) \mid t \in L_1 \cup L_2\} \enspace,
  \end{align*}
  which proves that $\tau_1 \cup \tau_2$ is quasi-alphabetic by
  Theorem~\ref{thm:Fact} (because $L_1 \cup L_2$ is recognizable
  by~\cite[Theorem~II.4.2]{gecste84}).
}

Let us move on to closure under intersection. For closure under
intersection, we would need to align the two input homomorphisms and
the two output homomorphisms at the same time and enforce equality
both-sided. The next theorem shows that we are not able to do this and
hence quasi-alphabetic relations are not closed under intersection.

\begin{theorem}
  \label{thm:Nonclosure}
  Any class~${\cal C}$ of tree transformations such that
  \[ \mathrm{lcssH} \subseteq {\cal C} \subseteq \calB(\mathrm{H},
  \Rec, \mathrm{lH}) \] is not closed under intersection.
\end{theorem}
\proof{
  Let $\Sigma = \{f/2, g/1, e/0 \}$. We consider the linear complete
  symbol-to-symbol homomorphisms $\psi_1, \psi_2 \colon T_\Sigma \to
  T_\Sigma$ that are defined by
  \begin{align*}
    \psi_1(f) &= f(x_1, x_2) & \psi_1(g) &=
    g(x_1) & \psi_1(e) &= e \\
    \psi_2(f) &= f(x_2, x_1) & \psi_2(g) &= g(x_1) & \psi_2(e) &= e
    \enspace.
  \end{align*}
  Clearly, $\psi_1$~and~$\psi_2$ belong to ${\cal C}$. Let us consider
  the context~$C = g(x_1)$ of~$C^1_\Sigma$. We observe that for every
  $m,n \geq 0$
  \begin{align*}
    f(g^m(e), g^n(e))\psi_1 &= f(C^m[e], C^n[e]) \\
    f(g^m(e), f^n(e))\psi_2 &= f(C^n[e], C^m[e]) \enspace.
  \end{align*}
  Let $L = \{f(g^m(e), g^n(e)) \mid m,n \geq 0 \}$. Clearly, $L$ is a
  recognizable language. Assume that there exists $\tau \in
  \calB(\mathrm{H}, \Rec, \mathrm{lH})$ such that $\tau = \psi_1 \cap
  \psi_2$. Since such bimorphisms preserve recognizable
  languages~\cite[Theorems II.4.2, II.4.16, II.4.18]{gecste84}, the
  image~$\tau(L)$ should be recognizable. But, $\tau(L)
  = \{ f(C^n[e], C^n[e]) \mid n \geq 0\}$, which is not
  recognizable. Hence no~$\tau$ with the given properties exists,
  which proves the statement.
}

\begin{corollary}[{\protect{of~Theorem~\ref{thm:Nonclosure}}}]
  ${\cal B}(\qH, \Rec, \qH)$ is not closed under intersection.
\end{corollary}

Finally, we note that ${\cal B}(\qH, \Rec, \qH)$ is trivially not
closed under complementation by Proposition~\ref{prop:Height}. Let us
consider now common operations on trees. We immediately observe that
intersection of a quasi-alphabetic relation with $\id_L$ where $L$ is
a recognizable language is again a quasi-alphabetic relation. Also the
union with~$\id_L$ is a quasi-alphabetic relation because $\id_L$ is a
quasi-alphabetic relation for every recognizable language~$L$ and
quasi-alphabetic relations are closed under union by
Corollary~\ref{cor:Union}.

In general, the tree transformations $\sub$~and~$\br$ (if we consider the branches as trees over an ranked
alphabet of figures of rank 0~and~1) are not quasi-alphabetic. Moreover, for $L \subseteq T_\Sigma(V)$
recognizable and $v \in V$, also the following relations $\tau$~and~$\rho$, which are defined for every $t \in
T_\Sigma(V)$ by $t\tau = t \bullet_v L$ and $t\rho = t /_v L$, are not quasi-alphabetic, in general
(cf.~\cite[Proposition~4.2 \& p.~191--200]{Bozapalidis92}). All these can easily be proved using
Proposition~\ref{prop:Height}. Moreover, in general, quasi-alphabetic relations are not closed under
$f$-top-concatenation (cf.~\cite[Proposition~3.6]{Bozapalidis92}).

Now, let us turn our attention to the translations computed by
quasi-alphabetic bimorphisms. In \cite{stetir07} it was shown that
they define the syntax-directed translations~\cite{ahoull72}. Here we
prove a more general result: the translations computed by
quasi-alphabetic tree bimorphisms are exactly the products of
context-free string languages (for definitions and details about
context-free string languages the reader is referred to
\cite[Section~I.6]{gecste84}).

\begin{theorem}[{\protect{cf.~\cite[Proposition~3.6]{Bozapalidis92}}}]
  For all context-free string languages $K_1$ and $K_2$ over the same
  alphabet $V$, there exists a quasi-alphabetic bimorphism~$B$ such
  that $\yd(\tau_B) = K_1 \times K_2$.
\end{theorem}
\proof{
  By~\cite[Corollary 2.4]{gecste84}, there exist recognizable tree
  languages $L_1 \subseteq T_\Sigma(V)$ and $L_2 \subseteq
  T_\Delta(V)$ such that $\{ \yd_V(t_1) \mid t_1 \in L_1\} = K_1$ and
  $\{\yd_V(t_2) \mid t_2 \in L_2\} = K_2$. Let $\phi \colon Y \to V$
  be a bijection, and $Y$ be disjoint with $\Sigma \cup \Delta$. Then
  extend~$\phi$ to $\phi_\Sigma \colon \Sigma \cup Y \to \Sigma \cup
  V$ and $\phi_\Delta \colon \Delta \cup Y \to \Delta \cup V$ such
  that $\phi_\Sigma|_\Sigma = \id_\Sigma$ and $\phi_\Delta|_\Delta =
  \id_\Delta$. We denote the ranked alphabets $\Sigma \cup
  Y$~and~$\Delta \cup Y$, in which all symbols of~$Y$ are nullary, by
  $\bar \Sigma$~and~$\bar \Delta$, respectively. Next, we define the
  ranked alphabet
  \[ \bar \Sigma \vee \bar \Delta = \{ \langle f, g \rangle \mid f \in
  \bar \Sigma, g \in \bar \Delta \} \] such that $\rk(\langle f, g
  \rangle) = \max(\rk(f), \rk(g))$. In a similar way the ranked
  alphabets $\Sigma \vee \bar \Delta$ and $\bar \Sigma \vee \Delta$
  are defined.  Without loss of generality, we can assume that $\bar
  \Sigma_0 \neq Y \neq \bar \Delta_0$ and $\Sigma_1 \neq \emptyset
  \neq \Delta_1$.

  Next we show how to embed a tree of~$T_\Sigma(V)$ into~$T_{\bar
    \Sigma \vee \bar \Delta}$. Roughly speaking, we read off the first
  components of the symbols of~$\bar \Sigma \vee \bar \Delta$ while
  neglecting the additional subtrees. However, we need to make sure
  that the neglected subtrees contain no symbols of~$Y$ because the
  quasi-alphabetic homomorphism cannot ignore the additional subtrees,
  but should clearly not produce a piece of output string for them. To
  this end, we define the linear top-down tree transducer~$M_\Sigma$
  with regular look-ahead $c$ such that $M_\Sigma = (\{\star\}, \bar
  \Sigma \vee \bar \Delta, \Sigma \cup V, \{\star\}, R)$, and for
  every $\langle f, g \rangle \in (\bar \Sigma \vee \bar \Delta)_k$ we
  have the rule
  \[ r = \star(\langle f, g \rangle(\seq x1k)) \to
  \phi_\Sigma(f)(\star(x_1), \dotsc, \star(x_{\rk(f)})) \] with
  look-ahead $c(r) = \langle f, g \rangle(\seq T1k)$ in $R$, where
  $T_1 = \dotsb = T_{\rk(f)} = T_{\bar \Sigma \vee \bar \Delta}$ and
  $T_{\rk(f)+1} = \dotsb = T_k = T_{\Sigma \vee \bar \Delta}$. In an
  analogous way the top-down tree transducer~$M_\Delta$ with regular
  look-ahead is defined. Let $L = \tau_{M_\Sigma}^{-1}(L_1) \cap
  \tau_{M_\Delta}^{-1}(L_2)$, which is recognizable by
  \cite[Corollary~IV.3.17 and Theorem~II.4.2]{gecste84}.  Next, we
  take the quasi-alphabetic homomorphism $\varphi \colon T_{\bar
    \Sigma \vee \bar \Delta} \to T_{\Sigma \vee \bar \Delta}(V)$,
  which is defined for every $\langle f, g \rangle \in (\bar \Sigma
  \vee \bar \Delta)_k$ by
  \[ \varphi_k(\langle f, g \rangle) = \begin{cases} \langle f, g
    \rangle(\seq
    x1m) & \text{if } f \in \Sigma_m \\
    \langle h_1,h_2 \rangle(\phi(f)) &
    \text{otherwise} \end{cases} \] where $\langle h_1,h_2 \rangle
  \in \Sigma_1 \times \Delta_1$ is arbitrary. Similarly, the
  quasi-alphabetic homomorphism ~$\psi \colon T_{\bar \Sigma \vee \bar
    \Delta} \to T_{\bar \Sigma \vee \Delta}(V)$ is defined. Now if we
  take the quasi-alphabetic bimorphism~$B = (\varphi, L, \psi)$, it
  should be clear that $\yd_V(t\varphi) = \yd_V(t\tau_{M_\Sigma})$ and
  $\yd_V(t\psi) = \yd_V(t\tau_{M_\Delta})$ for every $t \in T_{\bar
    \Sigma \vee \bar \Delta}$. Consequently, $\yd(\tau_B) = K_1 \times
  K_2$, which concludes our proof.
}

\section{Relation to Other Classes}
\label{sec:Relation}
In this section, we relate the class of quasi-alphabetic relations to
other known classes of tree transformations. We focus on classes of
transformations defined by bimorphisms~\cite{arndau82,dautis92,tata97}
and classes of transformations computed by various top-down tree
transducers~\cite{rou70,tha70,gecste84}. Clearly, every strictly
alphabetic (alphabetic in~\cite{gecste84}) homomorphism is
quasi-alphabetic and thus ${\cal B}(\saH, \Rec, \saH) \subseteq {\cal
  B}(\qH, \Rec, \qH)$. We start by showing that the class~$\text{\rm
  QREL}$ of transformations computed by finite-state
relabellings~\cite{eng75} is included in the class~${\cal B}(\saH,
\Rec, \saH)$.

\begin{proposition}
  \label{prop:QREL}
  $\text{\rm QREL} \subseteq {\cal B}(\saH, \Rec, \saH)$.
\end{proposition}
\proof{
  Let $\tau \in \text{\rm QREL}$. Since $\text{\rm QREL} \subseteq
  \text{\rm ln-TOP} = \text{\rm REL} \circ \text{\rm FTA} \circ
  \text{\rm lcH}$~\cite[Theorem~3.5]{eng75}, there exists a relabeling~$M$
  such that $\tau_M \subseteq T_\Sigma(V) \times T_\Gamma(Z)$, a
  recognizable tree language~$L \subseteq T_\Gamma(Z)$, and a linear
  and complete homomorphism $\psi \colon T_\Gamma(Z) \to T_\Delta(Y)$
  such that $\tau = \{ (t\tau_M^{-1}, t\psi) \mid t \in L\}$
  Moreover, by the constructions of~\cite{eng75}, $\psi$ is
  symbol-to-symbol and $\tau_M^{-1} \colon T_\Gamma(Z) \to T_\Sigma(V)$
  [i.e., $\tau_M^{-1}$ is computed by a deterministic
  relabeling]. Consequently, $\tau_M^{-1}$~and~$\psi$ are strictly
  alphabetic because every deterministic relabeling is strictly
  alphabetic. Thus, the strictly alphabetic bimorphism $(\tau_M^{-1}, L,
  \psi)$ defines~$\tau$.
}

The next proposition shows that every quasi-alphabetic relation can be
computed by a linear top-down tree transducer with finite
look-ahead~\cite{malgrahopkni07}. With that we establish rough lower
and upper bounds to the power of quasi-alphabetic bimorphisms.

\begin{proposition}
  \label{prop:TopFin}
  ${\cal B}(\qH, \Rec, \qH) \subseteq \text{\rm l-TOP}^{\text{\rm
      F}}$.
\end{proposition}
\proof{
  Let us consider a quasi-alphabetic bimorphism $B = (\vp, L, \psi)$ where $\vp
  \colon T_\Gamma(Z) \to T_\Sigma(V)$ and $\psi \colon T_\Gamma(Z) \to
  T_\Delta(Y)$. Without loss of generality, let $\vp$ be
  normalized. Moreover, let $N = (Q, \Gamma \cup Z, \Gamma \cup Z, I,
  R)$ be an fta recognizing~$L$. We construct the linear top-down tree
  transducer~$M$ with finite look-ahead~$c$ such that $M =
  (Q, \Sigma \cup V, \Delta \cup Y, I, R')$ and
  \begin{itemize}
  \item for every transition $q \to z \in R$ with $z \in Z$, we
    have the rule $r = q(z\vp) \to z\psi$ with look-ahead $c(r) =
    \{x_1\}$ in~$R'$, and
  \item for every transition $q \to f(\seq q1k) \in R$ with
    $f \in \Gamma_k$ and $\seq q1k \in Q$ we have the rule
    \[ r = q(\vp_k(f)(\varepsilon)(\seq x1n)) \to
    \psi_k(f)[q_1(x_{j_1}), \dotsc, q_k(x_{j_k})] \]
    with look-ahead $c(r) = \{ \vp_k(f) \}$ in~$R'$, where $j_i =
    \pos_{x_i}(\vp_k(f))$ for every $i \in [k]$.
  \end{itemize}
  First, let us prove $\tau_B \subseteq \tau_M$ by showing $q(t\vp)
  \Rightarrow_M^* t\psi$ for every $q \in Q$ and $t \in L(N)_q$. Let
  $t \in Z$. Then $q(t\vp) \Rightarrow_M t\psi$ using a rule
  constructed in the first item. Now let $t = f(\seq t1k)$ for
  some $f \in \Gamma_k$ and $\seq t1k \in T_\Gamma(Z)$. Moreover,
  let $\seq q1k \in Q$ be such that $t_i \in L(N)_{q_i}$ for every $i
  \in [k]$ and $q \to f(\seq q1k) \in R$. Then
  \begin{align*}
    q(f(\seq t1k)\vp) &= q(\vp_k(f)[t_1\vp, \dotsc, t_k\vp]) \\
    &= q(g(u_1[t_1\vp, \dotsc, t_k\vp], \dotsc, u_n[t_1\vp, \dotsc,
    t_k\vp]))
  \end{align*}
  where $\vp_k(f) = g(\seq u1n)$ for some $g \in
  \Sigma_n$ and $\seq u1n \in T_\Sigma(V)$. Let $j_i =
  \pos_{x_i}(\vp_k(f))$ for every $i \in [k]$. Then
  \[ q(f(\seq t1k)\vp) \Rightarrow_M
  \psi_k(f)[q_1(u_{j_1}[t_1\vp, \dotsc, t_k\vp]), \dotsc,
  q_k(u_{j_k}[t_1\vp, \dotsc, t_k\vp])] \]
  using a rule constructed in the second item. Note that the
  look-ahead restriction is trivially fulfilled. Clearly, $u_{j_i} =
  x_i$ for every $i \in [k]$ and thus we have
  \[ q(f(\seq t1k)\vp) \Rightarrow_M \psi_k(f)[q_1(t_1\vp),
  \dotsc, q_k(t_k\vp)] \enspace. \]
  By the induction hypothesis, we have $q_i(t_i\vp) \Rightarrow_M^*
  t_i\psi$ for every $i \in [k]$. Consequently, we obtain
  \[ q(t\vp) \Rightarrow_M \psi_k(f)[q_1(t_1\vp),
  \dotsc, q_k(t_k\vp)] \Rightarrow_M^* \psi_k(f)[t_1\psi, \dotsc,
  t_k\psi] = t\psi \enspace. \]
  This proves the auxiliary statement and $\tau_B \subseteq \tau_M$ if
  we consider states of~$I$.

  The converse inclusion can be proved using the statement: For every
  $q \in Q$, $t \in T_\Sigma(V)$, and $u \in T_\Delta(Y)$, if $q(t)
  \Rightarrow_M^* u$, then there exists $s \in L(N)_q$ such that $t =
  s\vp$ and $u = s\psi$. This can be proved by induction on the length
  of the derivation in~$M$. We omit the details here.
}

Next let us show that the class of alphabetic relations is essentially
different from the classes of transformations computed by top-down
tree transducers. For the specific class~$\text{\rm TOP}$ this was
already remarked in~\cite{Bozapalidis92} and here we only refine this
statement to the statements necessary for our purposes.

\begin{proposition}
  \label{prop:TopAlph}
  ${\cal B}(\aH, \Rec, \aH) \not\subseteq \text{\rm l-TOP}^{\text{\rm
      R}}$ and\/ $\text{\rm ln-TOP} \not\subseteq {\cal B}(\aH, \Rec,
  \aH)$.
\end{proposition}
\proof{
  It is known that $\text{\rm l-TOP}^{\text{\rm R}}$ equals $\text{\rm
    l-BOT}$, which is the class of all tree transformations computable
  by linear bottom-up tree transducers~\cite{tha73,eng75}. As claimed
  in~\cite[page~188]{Bozapalidis92}, the class ${\cal B}(\aH, \Rec, \aH)$ is
  incomparable to~$\text{\rm BOT}$, which is the class of all tree
  transformations computed by bottom-up tree
  transducers. Consequently, ${\cal B}(\aH, \Rec, \aH) \not\subseteq
  \text{\rm l-TOP}^{\text{\rm R}}$. Moreover, it is known that
  $\text{\rm lcH} \subseteq \text{\rm ln-TOP}$. Suppose that
  $\text{\rm lcH} \subseteq {\cal B}(\aH, \Rec, \aH)$. Then also every
  linear and complete inverse homomorphism can be implemented by an
  alphabetic bimorphism because alphabetic relations are trivially
  closed under inverses. However, the proof of the main theorem
  in~\cite[Section~3.4]{arndau82} then shows that alphabetic relations
  are not closed under composition. This
  contradicts~\cite[Theorem~5.2]{Bozapalidis92}, thus $\text{\rm lcH}
  \not\subseteq {\cal B}(\aH, \Rec, \aH)$. This yields $\text{\rm
    ln-TOP} \not\subseteq {\cal B}(\aH, \Rec, \aH)$.
}

Next we consider the relation of quasi-alphabetic and alphabetic
relations. We show that every quasi-alphabetic relation is also
alphabetic (ala~\cite{Bozapalidis92}). The strictness of this
inclusion can be obtained using Proposition~\ref{prop:TopAlph}.

\begin{theorem}
  \label{thm:QalphAlph}
  ${\cal B}(\qH, \Rec, \qH) \subseteq {\cal B}(\aH, \Rec, \aH)$.
\end{theorem}
\proof{
  Let us take a quasi-alphabetic tree bimorphism $B = (\vp, L, \psi)$ where $\vp
  \colon T_\Gamma(Z) \to T_\Sigma(V)$ and $\psi \colon T_\Gamma(Z) \to
  T_\Delta(Y)$. Without loss of generality, let $V \neq \emptyset \neq Y$. Let
  $\Sigma \vee \Delta$ be the ranked alphabet introduced
  in~\cite[Section~2]{Bozapalidis92}, and moreover let $v \in V$~and~$y
  \in Y$. We construct the linear tree homomorphism~$\rho \colon
  T_\Gamma(Z) \to T_{\Sigma \vee \Delta}(V \times Y)$ such that
  $\rho_Z(z) = \langle z\vp, z\psi \rangle$ for every $z \in Z$ and
  \[ \rho_k(f) = \langle t(\varepsilon)_w,
  u(\varepsilon)_{w'}\rangle(\seq x1k, \seq t1l) \]
  for every $f \in \Gamma_k$ where
  \begin{itemize}
  \item $t = \vp_k(f)$ and $u = \psi_k(f)$,
  \item $\{\seq i1m\} = \pos_V(t)$ and $\{\seq j1n\} = \pos_Y(u)$,
  \item $l = \max(m,n)$ and
    \[ t_a = \begin{cases}
      \langle t(i_a), u(j_a) \rangle & \text{if } a \leq \min(m,n) \\
      \langle t(i_a), y\rangle & \text{if } n < a \leq m \\
      \langle v, u(j_a) \rangle & \text{if } m < a \leq n
      \end{cases} \]
      for every $a \in [l]$, and
  \item $w = \word w1{k+m}$ and $w' = \seq{w'}1n$ are such that
    $t(w_a) =  \rho_k(f)(a)\pi_1$ for every $a \in [k+m]$ and $t(w'_b)
    =  \rho_k(f)(b)\pi_2$ for every $b \in [k+n]$ where
    $\pi_1$~and~$\pi_2$ are the usual projections to the first and
    second components, respectively, with $x\pi_1 = x = x\pi_2$ for
    every $x \in X$.
  \end{itemize}

  By~\cite[Theorem~II.4.16]{gecste84}, $\rho(L)$ is recognizable. An
  easy proof shows that
  \[ \tau_B = \{ t\vp_\Sigma, t\vp_\Delta)
  \mid t \in \rho(L)\} \] where $\vp_\Sigma$~and~$\vp_\Delta$ are the
  canonical alphabetic homomorphisms
  of~\cite[Section~2]{Bozapalidis92}. Hence, $\tau_B$ is an alphabetic
  relation by~\cite[Proposition~3.1]{Bozapalidis92}.
}

As an immediate consequence of Theorem~\ref{thm:QalphAlph}, we get the following result.

\begin{corollary}[{\protect{see~\cite[Proposition~3.7]{Bozapalidis92}}}]
  Quasi-alphabetic relations preserve the recognizable tree languages
  and the algebraic tree languages.
\end{corollary}

Finally, we need to show that linear top-down tree transducers are not sufficiently powerful to implement all
quasi-alphabetic relations.

\begin{proposition}
  \label{prop:qaln}
  ${\cal B}(\qH, \Rec, \qH) \not\subseteq \text{\rm l-TOP}$.
\end{proposition}
\proof{
  Let $\Sigma=\{ f/2, e/0 \}$ and $V =\{ v_1, v_2
  \}$. Moreover, let $\vp: T_\Sigma \to T_\Sigma(V)$ be a
  quasi-alphabetic tree homomorphism with $\vp_0(e) =
  f(v_1,v_2)$. Then $B = (\vp, \{e\}, \id)$ is a
  quasi-alphabetic tree bimorphism that defines $\{(f(v_1, v_2), e)
  \}$. It is known~\cite[Example~2.6]{eng75} that $\tau_B$ is not in
  $\text{\rm l-TOP}$, and hence ${\cal B}(\qH, \Rec, \qH)
  \not\subseteq \text{\rm l-TOP}$.
}

\begin{figure}
  \centering
  \includegraphics{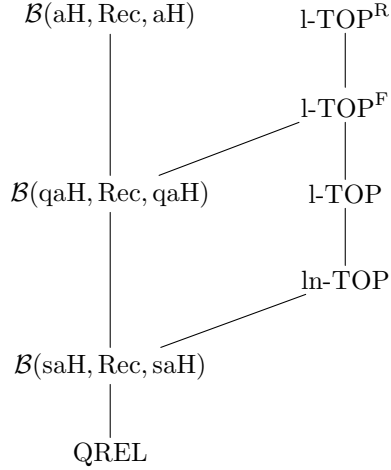}
  \caption{Hasse diagram.}
  \label{fig:Hasse}
\end{figure}

Let us collect our results in a \textsc{Hasse} diagram (see
Figure~\ref{fig:Hasse}). Note that in such a diagram every edge is
oriented upwards and denotes strict inclusion.

\begin{theorem}
  \label{thm:Hasse}
  Figure~\ref{fig:Hasse} is a \textsc{Hasse} diagram.
\end{theorem}
\proof{
  The following six statements are sufficient to prove the claim.
  \begin{align}
    \label{eq:Bim}
    \text{\rm QREL} &\subset {\cal B}(\text{\rm saH}, \Rec, \text{\rm
      saH}) \subseteq {\cal B}(\qH, \Rec, \qH) \subseteq {\cal B}(\aH,
    \Rec, \aH) \\
    \label{eq:Tops}
    {\cal B}(\text{\rm saH}, \Rec, \text{\rm saH}) &\subseteq
    \text{\rm ln-TOP} \subset \text{\rm l-TOP} \subseteq \text{\rm
      l-TOP}^{\text{\rm F}} \subseteq \text{l-TOP}^{\text{\rm R}} \\
    \label{eq:Quasi}
    {\cal B}(\qH, \Rec, \qH) &\subseteq \text{\rm l-TOP}^{\text{\rm
        F}} \\
    \label{eq:qHln}
    {\cal B}(\qH, \Rec, \qH) &\not\subseteq \text{\rm l-TOP} \\
    \label{eq:lnaH}
    \text{\rm ln-TOP} &\not\subseteq {\cal B}(\aH, \Rec, \aH) \\
    \label{eq:aHlR}
    {\cal B}(\aH, \Rec, \aH) &\not\subseteq \text{\rm
      l-TOP}^{\text{\rm R}}
  \end{align}
  Statement~\ref{eq:Bim} is mostly clear using
  Proposition~\ref{prop:QREL}. The strictness is due to the fact that
  $\text{\rm QREL}$ is closed under intersection whereas this
  is not true for ${\cal B}(\text{\rm saH}, \Rec, \text{\rm saH})$ by
  Theorem~\ref{thm:Nonclosure}. The final inclusion of~\eqref{eq:Bim}
  is proved in Theorem~\ref{thm:QalphAlph}. The inclusions of
  \eqref{eq:Tops} are all obvious and \eqref{eq:Quasi} is shown in
  Proposition~\ref{prop:TopFin}. Finally, the inequality~\eqref{eq:qHln}
  is proved in Proposition~\ref{prop:qaln} and inequalities
  \eqref{eq:lnaH}~and~\eqref{eq:aHlR} are proved in
  Proposition~\ref{prop:TopAlph}.
}

\pagebreak
\nonumsection{References}
\bibliography{quasi}
\bibliographystyle{ijfcs}

\end{document}